OmicsMapNet: Transforming omics data to take advantage of Deep Convolutional Neural Network for discovery

Shiyong Ma[1], Zhen Zhang[1,2,*]

Center for Biomarker Discovery and Translation, Departments of Pathology[1] and Oncology[2] Johns Hopkins Medical Institutions 1550 Orleans Street, Baltimore, MD 21231, USA

*corresponding author.

**We developed OmicsMapNet to take advantage of existing deep leaning frameworks to analyze high-dimensional omics data as 2-dimensional images. The omics data of individual samples were first rearranged into 2D images in which molecular features related in functions, ontologies, or other relationships were organized in spatially adjacent and patterned locations. Deep learning neural networks were trained to classify the images. Molecular features informative of classes of different phenotypes were subsequently identified. As a proof of concept, we used the KEGG BRITE database to rearrange RNA-Seq expression data of TCGA diffuse glioma samples as treemaps to capture the functional hierarchical structure of genes in 2D images. Deep Convolutional Neural Networks (DCNN) using tools from TensorFlow was able to learn to predict the grade of TCGA LGG and GBM samples with relatively high accuracy. The most contributory features extracted from the trained DCNN were further checked in pathway analysis for their plausible functional involvement. Compared to the typical global significance test-based approaches of supervised discovery analysis of omics data, OmicsMapNet aggregates related molecular features together to improve their collective contrast between phenotypes while at the same time harvesting the power of deep learning frameworks and tools developed for image object recognition to extract such features.**

In recent decades, high throughput proteogenomic analysis technologies have become much more mature(Nesvizhskii 2014; Mertins et al. 2016; H. Zhang et al. 2016; Bing Zhang et al. 2019). This has led to an explosion of genomics, transcriptomics, proteomics and other datasets from large-scale profiling efforts of biological/clinical samples. The analysis of genes and the products of genes contributes to deeper understandings of human diseases at molecular level. Molecular biological omics data are now serving as important resources in biomarkers research for initial discovery, further distillation, and possible corroborative verification(Ludwig and Weinstein 2005; Skates et al. 2013; Bai Zhang, Tian, and Zhang 2014; Wishart 2016; Thomas et al. 2016; Geyer et al. 2017; A. Srivastava and Creek 2018).

Until recently, much of the omics data analysis for biomarker discovery involves some form of univariate global significance tests with adjustment/control for multiple-testing (**Fig. 1a**). It is often only until the number of potential targets has been significantly reduced that more complex multivariate approaches, such as GSEA(Mootha et al. 2003; Subramanian et al. 2005), differential dependency network (DDN) analysis(Bai Zhang et al. 2009), MCCA(Witten and Tibshirani 2009), CAMERA(Wu and Smyth 2012), GSVA(Hänzelmann, Castelo, and Guinney 2013), Enrichr(E. Y. Chen et al. 2013; Maxim V. Kuleshov et al. 2016) and DIABLO(Singh et al. 2016; Rohart et al. 2017) are used to assess their performance in combination. However, as we have learned from mutational landscape analysis of cancers, with the exception a few gene "mountains" mutated at high frequency, the majority of early genomic alterations happen at low frequency (Jones et al. 2010; Williams et al. 2016), which makes it difficult to identify such early changes through a globally applied significance test. It is desirable therefore to develop supervised multivariate approaches that are capable of applying existing knowledge to aggregate "significances" from functionally related features that are yet sparsely scattered in high-dimensional omics data for the purpose of discovering molecular changes that would otherwise be missed.

The use of existing knowledge in high-dimensional omics data analysis have been previously reported(Wang et al. 2013; McDermott et al. 2013; Liu et al. 2016; Wang, Vasaikar, et al. 2017; Wang, Ma, et al. 2017). A major difficulty is to numerically code such knowledge to allow for efficient integration in computational analysis. Rapid advances in machine learning frameworks have been able to provide ready-to-use and well-tested tools for the application of deep neural network for visual object recognition using image data. The learned neural networks, in turn, can also provide information on visual features in the images that are most contributory to object classification. In our current approach, we rearrange omics expression data as 2-dimensional (2D) images in which molecular features related in functions, ontologies, or other relationships were organized in spatially adjacent and patterned locations (**Fig. 1b)**. We then take advantage of existing deep leaning frameworks to train deep learning neural networks to classify the images into known phenotype classes. The identified informative features for the trained neural network provides the basis for discovery of phenotype specific biomarkers.

For this research, as a proof of concept, we used treemap (Shneiderman 1992) to rearrange omics data of individual samples into hierarchically structured 2D images based on knowledge of hierarchical mapping and functional annotation of genes extracted from the KEGG BRITE reference database that captures the functional hierarchies of biological objects – especially the KEGG objects (http://www.kegg.jp/) (M. Kanehisa 2006; Minoru Kanehisa et al. 2012, 2008, 2016).

Treemap is a space-filling method for visualizing large scale hierarchical structures(Shneiderman 1992; Bederson, Shneiderman, and Wattenberg 2002). The treemap methods has previously been used to visualize omics expression data. For example, Baehrecke applied treemaps to visualize the microarray based gene expression data in the context of GO hierarchical annotations(Baehrecke et al. 2004), Halligan used treemaps to display quantitative proteomics data using simplified version of GO annotation - Generic GO slim(Halligan et al. 2007), and Bernhardt(Bernhardt et al. 2009) visualized gene expression data via Voronoi treemaps using GO slim generic and KEGG Orthology. In our current work, we used a similar process to construct treemaps that are conducive to deep neural network learning. It is possible to employ other methods to convert omics expression data into 2-dimensional images, including algorithms that allocate spatially genes as points in a 2D plane based on some types of "closeness" measures (e.g., correlations).

For the current work, to test our approach, we explored a gene expression dataset from the Cancer Genome Atlas (TCGA) to create treemap images. We obtained the gene expression matrix of TCGA diffuse glioma samples from Ceccarelli's paper (Ceccarelli et al. 2016). The grading of gliomas is important as erroneous grading may lead to either over- or under-treatment, affecting particularly the radiotherapy dosage and the use of chemotherapy. The World Health Organization (WHO) grading of gliomas was traditionally based on histologic type and malignancy grade (Louis et al. 2007), and since 2016 the WHO grading takes several additional molecular features (IDH-mutant, 1p/19q-codeleted, etc.) into consideration to define several new gliomas(Louis et al. 2016).

Convolutional Neural Networks (CNN) was inspired by visual cortex of the brain, and are designed to process 2D images (Krizhevsky, Sutskever, and Geoffrey E. 2012). Deep CNN was used in our work to learn the converted omics 2D images and to extract molecular features that differentiate sample phenotypes. Significant commercial and community efforts have led to the availability of a large number of deep learning frameworks that provide building blocks for efficient and effective development of deep learning neural networks (Jia et al. 2014; Tokui et al. 2015; Chollet 2015; Abadi et al. 2016; Paszke et al. 2017). In this work, the CNN was implemented in Python employing the TensorFlow framework (Abadi et al. 2016). Specifically, CNN was trained to learn to predict the grade of the TCGA LGG and GBM samples, the contributory genes for the classification were identified from the feature maps of the trained CNN.

To demonstrate the effectiveness of OmicsMapNet approach, comparative analyses were performed. The accuracies of tumor grade prediction were compared against those using logistic regression and gradient boosting decision trees on the gene expression data without 2D treemap transformation. The differential expressed genes from traditional global significance-test was obtained, pathway analysis using the contributory genes extracted from the feature maps further helps to identify plausible functional pathways that are yet not discovered in differential expressed genes.

**Results**

*Transforming TCGA LGG&GBM gene expression data*

In OmicsMapNet, a four-layer hierarchical structure of functional annotations was extracted from KEGG BRITE hierarchy files (see Methods). The genes were then assigned to the child nodes of corresponding functional annotations in the hierarchical tree, and a fifth layer was generated. The treemap images were created based on the five-layer hierarchical tree structure.

To create treemap images, we used the TCGA LGG&GBM dataset from Ceccarelli's paper (Ceccarelli et al. 2016). 20330 genes were initially obtained from the gene expression matrix, and 17715 genes were obtained after eliminating genes with extremely low expressions. The 17715 filtered genes were mapped to the KEGG IDs, and 7095 genes could be processed by OmicsMapNet to generate the structure (spatial allocation of genes) of the treemap image which contains 10772 gene squares (**Fig. 2a-d**). In the treemap image, each rectangle represents one gene, the pseudo-color represents the normalized intensity of the gene. In the dataset, RNA-Seq analysis was performed on 667 samples of which 607 were labelled with WHO grade (Louis et al. 2016, 2007). In total, therefore, 607 treemap images were created using the same fixed treemap structure.

*Learning and predicting the grade of tumor samples using DCNN*

A Deep Convolutional Neural Network (DCNN) with 3 convolutional layers and 2 fully connected layers was further developed (**Fig. 3a**). Details are provided in the Methods. The DCNN was trained to learn the grade of the tumor samples with the generated treemap images of the TCGA LGG&GBM samples as input and the corresponding sample WHO grades as output.

Detailed clinicopathological characterization and other information were provided in the original report (Ceccarelli et al. 2016). For the current work, the distribution of subjects in WHO grade II, III, and IV for 607 TCGA LGG&GBM samples were 215, 239, and 153, respectively.

The performance of the DCNN was assessed using 10-fold stratified cross validation. The resulting mean and median accuracies were 75.09% (95% CI: 70.38 – 79.79%) and 74.35%, respectively. In comparison, when the same model design and learning procedure applied to label-permutated samples, the mean and median accuracies were 37.72% (95% CI: 35.19 – 40.24%) and 38.85%, respectively. As shown in **Fig. 3b**, the difference was extremely significant (Wilcoxon test: p-value = 1.77e-3).

From the receiver-operating-characterization (ROC) curves (**Fig. 3c-e**), G2 and G3 are relatively more difficult to be distinguished, the mean area-under-curve (AUC) of G2 is 0.86 and the mean AUC of G3 is 0.83. Contrarily, the mean AUC of G4 is 0.99, which indicates that G4 can be distinguished from G2 and G3 with a higher accuracy.

*Identifying contributory genes from feature maps*

With the relatively high prediction accuracy of the DCNN model, we were able to use the deep learning model to identify a set of genes that are highly contributory to the classification. Feature maps were output from the convolutional layers and/or the following pooling layers, and they present the strengths and spatial allocations of activations (He et al. 2015). It have been found that the strong activation points in higher convolutional layer feature maps have latent semantic content (He et al. 2015; Zeiler and Fergus 2014).

We found that the sum of pixel intensity of feature maps reduced from low convolutional layer to high convolutional layer. In Pool3, most of the features maps had sparse nonzero pixel intensities, the pixel intensities were concentrated to several feature maps. Even in the strongest feature maps (the maps whose sum of pixel intensities were the highest), the nonzero pixels were likely to be sparse (**Fig. 4a-c**). In this work, considering Pool3 layer is directly linked to the downstream fully connected layer, the feature maps generated from Pool3 layer were used for the gene set selection. To be specific, Conv3 layer has 64 channels, so 64 feature maps were outputted from the following Pool3 layer, and we used the strongest feature map for contributory gene selection. We first collected the strong response pixel points in all the selected feature maps, and obtained the positions and selection times of the strong responses (see Methods). The positions of the genes in the treemaps were saved (Supplementary File 1_TreemapStructure_GenePositionsGeneID.xlsx). A coarse projection method (see Methods) was then adopted to map the selected strong response pixels to gene IDs. The corresponding genes and number of selection of those genes were obtained (Supplementary File 2_SummaryofGenes.xlsx).

*Analysis of obtained contributory genes and comparison with known key genes from literature*

From the obtained gene list (Supplementary File 2_SummaryofGenes.xlsx) ), genes with corresponding gene name NEUROD1, FOSB, HUF4, MESP1 and ko03000"Transcription factors" functional annotation were selected 456 times in the 607 samples, and they are assumed to be associated with tumor grading from the model. Additionally, number of individual genes could each appears in multiple places in the treemap image due to possible multiple KEGG functional annotations. Based on the local connectivity property of DCNN model, it was possible that a gene might be selected in one functional position yet not in its other functional positions. In brief, genes are selected locally, and the selection are affected by spatial neighbor genes either within the functional groups or allocated in the neighbor functional units. For example, for the gene CLIC1, the pixel point with ko04040 "Ion Channels: Chloride channels" annotation term was selected only 8 times, comparatively, the pixel point with ko04147 "Exosome: Exosomal proteins" annotation term was selected 313 times.

We were further interested in comparing the obtained contributory gene list with some known keys genes reported in literature (Ceccarelli et al. 2016; Zheng et al. 2008; Eckel-Passow et al. 2015) that are associated with the development and progression of central nerve system tumor(**Supplementary Table 1**). We observed that the number of selections for these genes varied significantly. For example, IDH2 was selected 330 times, this result is consistent with the grading method of gliomas from WHO. However, CIC was only selected for 58 times. Again, the number of selections for the same gene might be different in different functional units. For gene ARID2, it was selected 10 times in "ko03036 Chromosome" "Eukaryotic Type" unit, and 25 times in "ko03021 Transcription machinery" "Eukaryotic type" unit. Also for gene SETD2, it was selected 1 times in "ko01000 Enzymes" "2. Transferases" unit, and was selected 5 times in "ko03036 Chromosome" "Eukaryotic Type" unit.

*Pathway analyses of the contributory genes from deep learning*

From the contributory gene list, NEUROD1 gene is among the top selected ones, it has been reported that the developmental transcription factor NEUROD1 is found to be abnormally expressed in a subset of aggressive neuroendocrine tumors (Osborne et al. 2013), and also it is found that the homozygous mutations in NEUROD1 are responsible for neurological abnormalities (Rubio-Cabezas et al. 2010). The abnormal expression of FOSB gene was found to be frequent in breast carcinomas (Milde-Langosch et al. 2003) and pancreatic cancer (Kim et al. 2010). Numbers of highly selected genes could be further investigated to check their related biological functions and clinical impacts.

The top 320 most selected genes (appears in over half of the samples) from OmicsMapNet method were further investigated using REACTOME pathway analysis (**Fig.5**). It is worth noting that pathway of "Signaling by Overexpressed Wild-Type EGFR in Cancer" was found significantly enriched (p value with Bonferroni step down = 4.3e-2), the abnormal regulation of EGFR pathway has been known to be correlated with malignant tumors of central nervous system (Heimberger et al. 2002). Also, the pathway of "Mitochondrial ABC transporters" was found significantly enriched (p value with Bonferroni step down = 1.4e-4), this pathway was also reported to be associated with tumor progression and clinical outcomes (Fletcher et al. 2010).

*Discrepancies of analysis results between OmicsMapNet and global univariate test method*

Typical global univariate test was performed for comparisons (see Methods). First, the Principle Component Analysis on the RNA seq data (Fig.6a) shows that G4 is easier to be differentiated, contrarily, G3 and G2 are relatively more difficult to be differentiated. Second, the hierarchical clustering analysis of the top 500 most variable genes (Fig.6b) also show that G4 is relatively easier to be separated from the other grades. We then checked the differential expressed genes in different grades via pairwise comparisons, and the volcano plots are shown in Fig.6c. The contributory genes selected by OmicsMapNet approach were further plotted in the volcano figures (the red dots). It can be noticed that the many selected contributory genes do not have large fold change and the p-values of some genes are not small enough to be statistically significant. Only a limited small number of genes are overlapped between contributory genes and the differentially expressed genes. Particularly for grade 2 and grade 3, the Venn Graph (Fig.6d) shows there are only 4 overlapping genes.

***Classifying grade 2 and grade 3 samples using OmicsMapNet, Logistic Regression and Gradient Boosting Decision Trees***

Previous analyses in this paper have shown the difficulties to accurately classify grade 2 and grade 3 samples, in this section, we compared the grade 2 vs grade 3 classification performances between OmicsMapNet and state-of-the-art methods. To this end, we used logistic regression and gradient boosting decision trees with XGBoost (T. Chen and Guestrin 2016) as the benchmarks.

For OmicsMapNet, the DCNN architecture and training procedure was kept the same as that in previous grade classification, and 10 stratified cross validation was also used. The performance is shown in Fig.7a, and the mean of AUC is 0.86. When the same classification procedure was performed using Logistic Regression on this RNA-Seq expression data, the mean of AUC is 0.79 (Fig.7b). And for the state-of-the-art method gradient boosting decision trees which is currently one of the most popular machine learning models, the mean of AUC is 0.72 (Fig.7c).

Considering the input dimension is high compared to the number of samples (Grade 2: 215, Grade 3: 239), to reduce overfitting, we sampled limited number of genes from the differential expressed gene set and re-used Logistic Regression and gradient boosting decision trees (GBDT) to do classification on the sampled sub-set of genes. For each gene selection number, we sampled the given number of genes from the differential expressed gene set for 50 times, the 10 fold stratified cross validation was then adopted to measure the performances, the means and standard deviations of the AUC was plotted against the gene selection number. All the results indicate that OmicsMapNet achieves a more accurate classification for the grade 2 and grade 3 samples, compared to other benchmark algorithms.

**Discussion**

Omics expression dataset (e.g MS proteomics data, mRNA expression data) usually contains thousands of genes or proteins, the dimensions are much larger than the number of samples. To discover effective biomarkers for certain phenotypes or clinical properties, feature selection from the high dimensional data is often required. In recent years, deep learning is proved to have strong capacity in discovering intricate patterns in high-dimensional data(Lecun, Bengio, and Hinton 2015), with this consideration, we developed OmicsMapNet. The aim is to convert the high-dimensional omics data into a format on which deep learning can be used for feature discovery.

Specifically, since DCNN has been widely used for visual understanding with success (Ranzato et al. 2006; Szegedy et al. 2015; Simonyan and Zisserman 2015), in this current work, we transformed omics expression data into 2-dimensional images and developed the deep CNN model to understand the grade of TCGA LGG and GBM samples. Traditional approaches rely heavily on global univariate analysis, and ignore the correlations between different genes at the initial discovery step. Yet, the DCNN pooled the original treemap images. During spatial pooling method, the outputs of several nearby feature detectors are combined into local or global "bag of features", the aim is to enable compact representations, better robustness to noise(Boureau, Ponce, and LeCun 2010). Therefore, in this work, the DCNN provides a way to jointly considering multiple variant locally with similar function annotations. Typical global univariate test was performed for comparison, large discrepancies exist between the analysis results, this indicates that OmicsMapNet approach could probably be a useful method to provide supplementary patterns of the omics data and discover important phenotype associate genes and pathways that are missed by typical global univariate test.

For classifying grade 2 and grade 3 samples, OmicsMapNet achieved a better performance in terms of the AUC values compared to Logistic Regression and GBDT. A possible explanation could be that the transformation of original omics data aggregates functional related genes together and makes the phenotype related information in the omics data more explicit. To test this particular hypothesis, further investigations need to be done and yet it is out of the scope of this current paper.

Finally, although the RNA-Seq expression data was used to generate the treemap images in this work, yet, OmicsMapNet method can also be used for protein expression data and other types of quantitative omics and clinical data.

**Methods and technical details**

*Omics expression data*

The RNA-Seq expression matrix of the TCGA LGG&GBM dataset was obtained at https://tcga-data.nci.nih.gov/docs/publications/lgggbm_2015/LGG-GBM.gene_expression.normalized.txt. To be specific, we first used the R package edgeR(Robinson, McCarthy, and Smyth 2010) to normalize the RAW count number into TMM value (log2)(Risso et al. 2014). We filtered out the genes with extremely low TMM values. The threshold of TMM value that we used for filtration is -5. For the remaining genes, gene names were matched to KEGG IDs using R package UniProt.ws from Bioconductor. For KEGG-ID which can be mapped to multiple genes in the data matrix, we chose the gene with the highest mean expression value.

*Converting Omics expression data into treemap images*

To convert omics expression data into treemap images, we first parsed the KEGG BRITE structure at http://rest.kegg.jp/get/br:br08902. In this work, we only considered the Genes and Proteins branch of the BRITE hierarchical structure. Therefore, a three-layer hierarchical tree with 57 child nodes was initially generated, and each child node represents a ko hierarchy file. Then we explored the ko hierarchy files and extracted the next layer of functional groupings from those files and updated the generated tree with a third layer, the finer tree contains 234 child nodes. Then we further assigned the genes to the corresponding child nodes of the tree structure based on the KEGG IDs. Finally, a five-layer hierarchical tree was generated. Due to the fact that one gene might have multiple KEGG functional annotation, these genes appear in several positions in the tree.

Second, to spatially arrange the genes of a sample into a 2-dimensional image, rectangle treemaps were used. Rectangle units were used to represent genes in the treemaps. Considering that each rectangle unit represents one gene, each rectangle unit was assigned the same size of area. To put the rectangle units into the treemap, several algorithms have been proposed previously. In this paper, the pivot method (Bederson, Shneiderman, and Wattenberg 2002) was used. To realize this, the R package TreeMap from CRAN was used and modified. Furthermore, within each functional group, the genes were sorted according to the median value of the TMM (log2) values across all the samples. All the treemap images have exactly the same spatial arrangement of the genes.

Third, we colored the treemaps based on the normalized value of the gene expression abundances. The normalized gene expression values were mapped to colors in the BlueYellwRed heatmap sample by sample. For each sample, the highest TMM value (log2) was mapped to the Red color, and the lowest TMM value (log2) was mapped to the Blue color, 256 colors were interpolated, TMM values (log2) were linearly mapped to the corresponding colors in the BlueYellwRed heatmap.

The created original treemap images contains 1024*1024 pixels, the images were subsampled to 512*512 pixels before input to DCNN.

*The designed deep Convolutional Neural Network*

The architecture of the designed deep CNN is with 3 convolutional layers (conv), 3 pooling layers and 2 fully connected layers (fc), the RELU(Tang 2013) activation function was used in Conv1-3 layers and Fc1 layer. The softmax activation function was used for the output layer.

Adam Optimizer(Kingma and Ba 2014) was used as the optimizer, and the loss function is Cross Entropy. Conventional convolution neural network layer normally contains alternating CNN layers and pooling layers, this method was adopted in this work.  Max Pooling method was used as the pooling function. For the conv layers, the filter size is 3, the first and second conv layer has 32 filters respectively, and the third conv layer has 64 filters, the strides of all the conv filters are 1, and the strides for all the pooling layers are 2, the padding method adopted is "VALID". The first Fc layer and second Fc layer have 128 and 3 neurons respectively. For other hyper-parameters of the model, dropout method which was proposed by Hinton(N. Srivastava et al. 2014) was used on Fc1 layer (keep probability is 0.75), L2 regularization was used for both Fc layers to further reduce overfitting and the beta parameter is 0.01, learning rate is 0.001, batch size for learning is 29, the number of iterations is 300 and early stopping number is 10. 10-fold stratified cross validation method was adopted.

*Selection of contributory gene set based on the feature maps of CNN*

For each sample, the Pool3 feature map with the largest sum of pixel intensities was selected. And to summarize the strong response points of the 607 selected feature maps, the top 10% strongest points in each feature maps were chosen. Then we calculated the number of selections of each pixels in the feature maps of all samples.

In computer vision, for object detections, by projecting the strong responses in feature maps back to the original image(He et al. 2015), features could be possibly observed and therefore discovered. In this work, we adopted a coarse method to project the strong responses to original treemap. Due to the pooling operation, the feature map is down sampled from original treemap. The size of feature map from Pool3 is 62*62, so each pixel in the Pool3 feature map was linearly projected to a square area with PR*PR pixels in the treemap image (PR = (1024/62)), the projection is shown in **Supplementary Figure 1.** If the center of a gene's rectangle is within the square area of selected response point, it is assumed that the selected response point can be projected to the gene.

*Pathway analyses of selected gene set*

To further explore the selected contributory genes, Cytoscape(Shannon et al. 2003) ClueGO(Bindea et al. 2009) was used. In this work, REACTOME pathway(Fabregat et al. 2016) analyses were done. For the pathway analysis, we chose the specific terms, and pathways only with Bonferroni step down corrected p-value smaller than 0.05 were shown.

*Differential gene expression analysis using global univariate test*

To conduct differential expression analysis of this RNA-seq data, we used the R package edgeR. The raw count was first normalized to counts-per-million (CPM) value. For this dataset, we chose to retain genes if they were expressed at CPM above 0.5 in at least two samples. The CPM value was further log2 transformed. Principle component analysis was done on the normalized log2 CPM values of the retained genes. To reduce computation burden, the 500 most variable genes were selected for the hierarchical clustering analysis. The result was plotted using heatmap. For further differential expression analysis, raw count data was normalized and transformed using voom (Law et al. 2014), the differential expression analyses were then conducted between different grades using linear model.

*Using Logistic Regression and Gradient Boosting Decision Trees to classify grade 2 and grade 3 samples*

For logistic regression, we used LogisticRegressionCV function from scikit-learn python package. For the dataset$\{(X_i, y_i)\}$, the cost function is

$$C \sum_{i=1}^{n} \log(\exp\left(-y_i(X_i^T \omega + c)\right) + 1) + \min_{\omega, c} \frac{1}{2} \omega^T \omega \qquad (1)$$

L2 norm were used to regularize the model, "liblinear" optimizer was used to minimize the cost, a build-in 10 fold cross validation was applied to find the optimal hyper-parameters C.

For GBDT, XGBoost was used for implementation. For the tree $f(x)$ as

$$f_t(x) = \omega_{q(x)}, \omega \in R^T, q: R^d \to \{1,2,\ldots,T\} \quad (2)$$

$\omega$ is the vector of scores on leaves, $q$ is a function assigning each data point to the corresponding leaf, and $T$ is the number of the leaves, the complexity of the tree is defined as

$$\Omega(f) = \gamma T + \frac{1}{2}\lambda \sum_{j=1}^{T} \omega_j^2 \quad (3)$$

We added L2 norm regularization by setting $\gamma = 0, \lambda = 1$, early stopping is 10, and max depth to 3.

For both of the two models, other hyper-parameters were tuned empirically within typical values or set as default to avoid large amount computation of hyper-parameter tuning.

ACKNOWLEDGMENTS

This research was partially supported by US National Institutes of Health (NIH) grants U01CA200469 and U24CA210985 to Z.Z.

AUTHOR CONTRIBUTIONS

Z.Z. designed the study. S.M. developed the method and analyzed the results. Z.Z. supervised the study. S.M and Z.Z wrote the paper.

COMPETING INTERESTS

The authors declare no competing financial interest.

**Figure legends:**

Figure 1 | Outline of OmicsMapNet Workflow compared to traditional approach (a) Traditional approaches typically rely on global univariate significant test for initial feature discovery, the correlations between different genes are not considered at this step. The selected genes are then analyzed using multivariate approaches. (b) OmicsMapNet method transforms omics expression data to 2-dimensional treemap images. Deep learning neural networks are then trained to classify the images, genes are selected in a supervised method.

Figure 2 | The hierarchical structure is shown layer by layer (a)-(d). Each gene rectangle (d) represent one gene, the gene rectangles have the same size of area. Therefore, the size of the rectangle of functional unit represents the number of genes within the unit. The colors of the gene rectangle(d) represent the normalized gene abundances, and the colors of the functional unit rectangles(a)(b)(c) were randomly generated.

Figure 3 | (a) The architecture of DCNN. It contains 3 convolutional layers (Conv), 2 fully connected layers (Fc). (b) The prediction accuracies of 10 folds (Original samples vs Label permutated samples). (c) ROC curves of the G-II, the mean AUC is 0.86. (d) ROC curves of the G-III, the mean AUC is 0.83. (e) ROC curves of the G-IV, the AUC is 0.99.

Figure 4 | Example of feature maps from Pool3. Since signal intensity of Pool3 feature map is low, heatmaps were used to visualize the feature maps. It could be noticed that most activations in the feature map is 0. (a) Top three strongest feature maps of Sample TCGA-HT-7467 (G-II). (b) Top three strongest feature maps of Sample TCGA-HT-A619 (G-III). (c) Top three strongest feature maps of Sample TCGA-19-2620 (G-IV).

Figure 5 | Results of REACTOME pathway enrichment analysis on selected genes, enriched pathways terms and the associated genes are presented. The depicted pathway terms were significantly enriched with Bonferroni step down corrected pvalues < 0.05. (Size of the circle represents the significance degree, larger circle size means lower p value.)

Figure 6 | (a) Principle Component Analysis of the RNA seq data. (b) hierarchical clustering analysis of the top 500 most variable genes. (c) Volcano plot of differential expressed genes via pairwise comparative analysis. (d) Venn Graph of differential expressed genes (Grade 2 and Grade 3) and contributory genes selected by deep learning.

Figure 7| (a) Grade2 and grade 3 sample classification results of OmicsMapNet (b) Grade2 and grade 3 sample classification results using Logistic regression (c) Grade2 and grade 3 sample classification results using Gradient Boosting Decision Trees (XGBoost) (e) Grade2 and grade 3 sample classification results using Logistic regression and GBDT on sampled subset of differential expression genes.

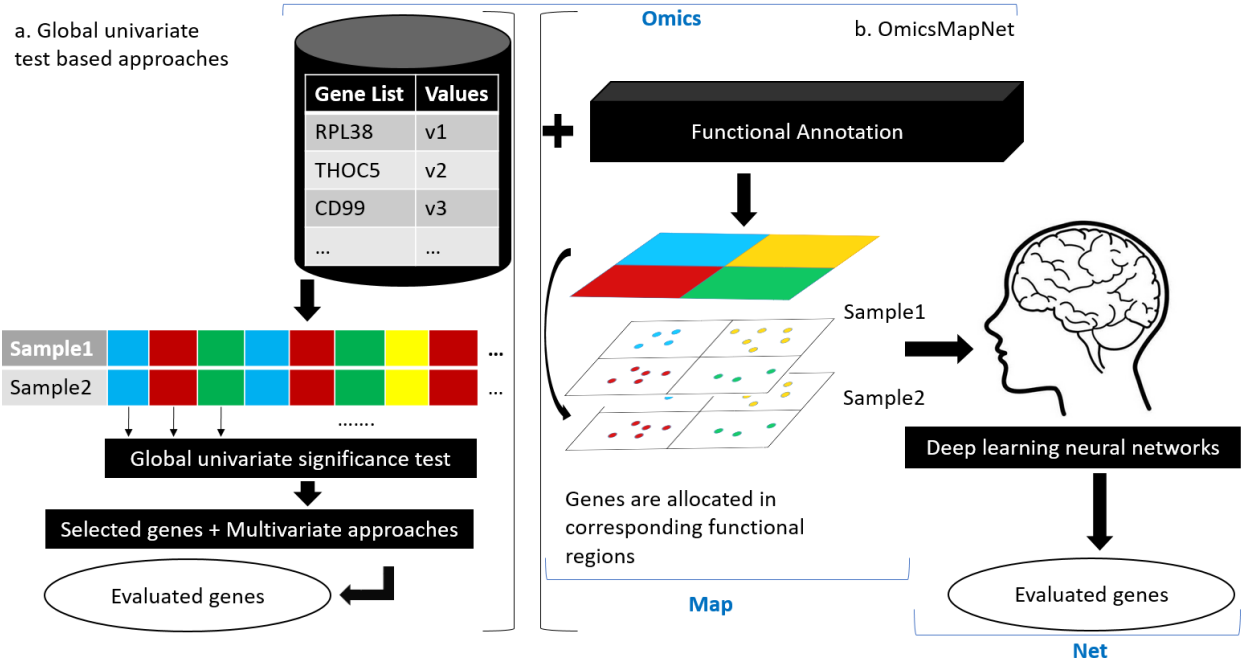

Figure 1

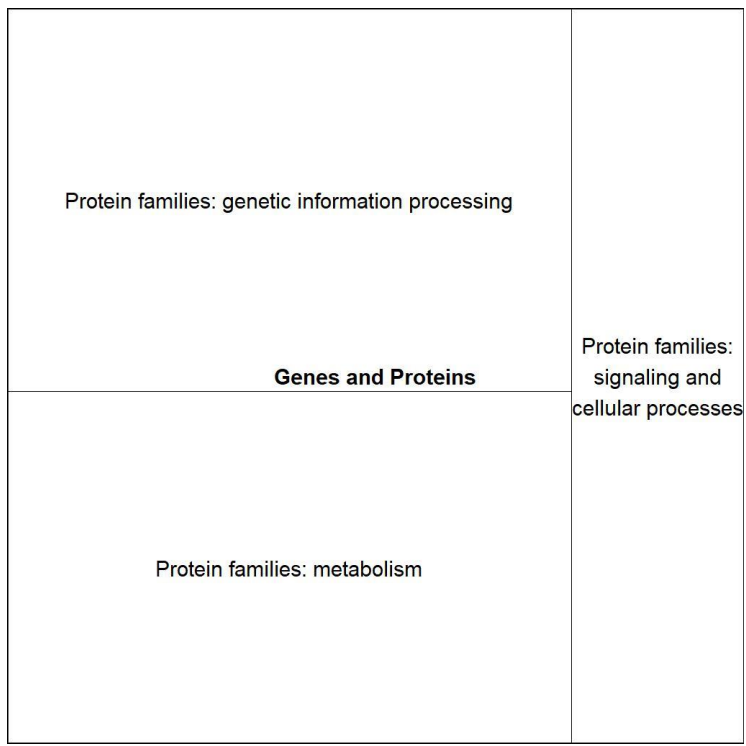

Figure 2 (a)

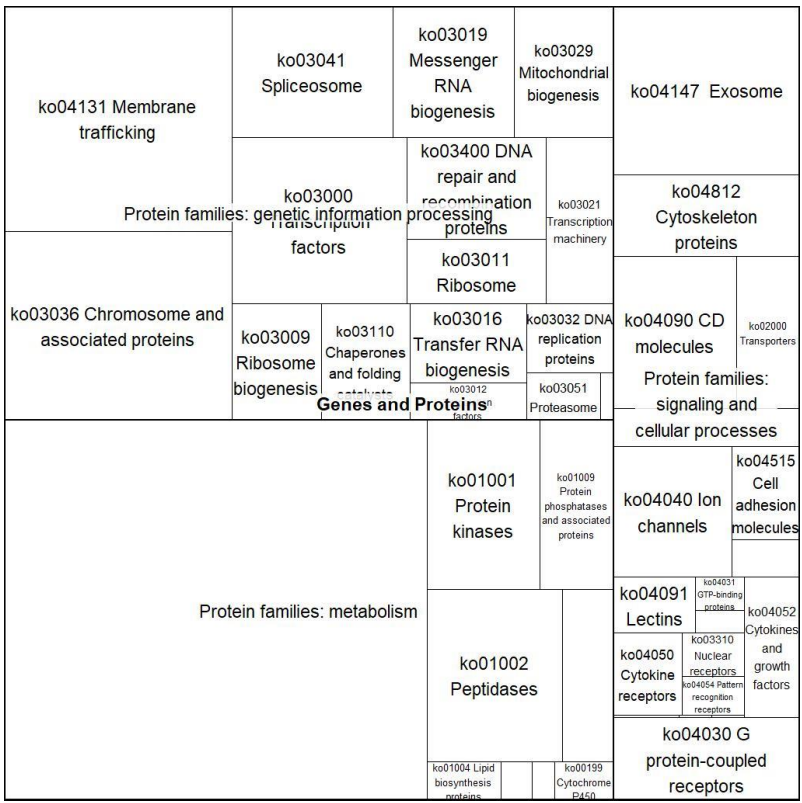

Figure 2 (b)

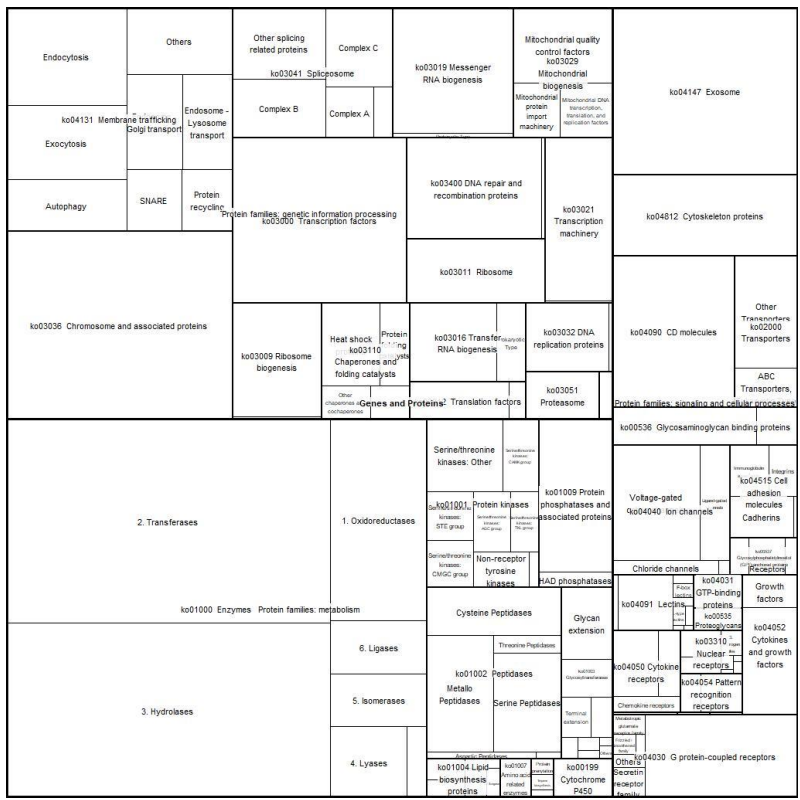

Figure 2 (c)

Figure 2 (d)

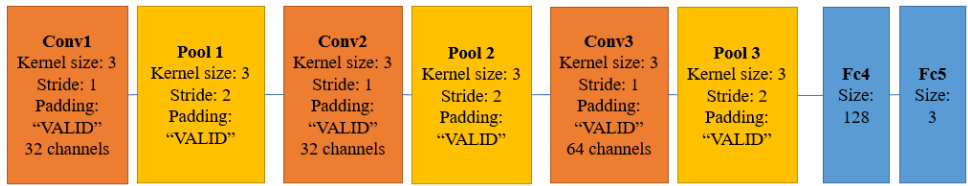

Figure 3 (a)

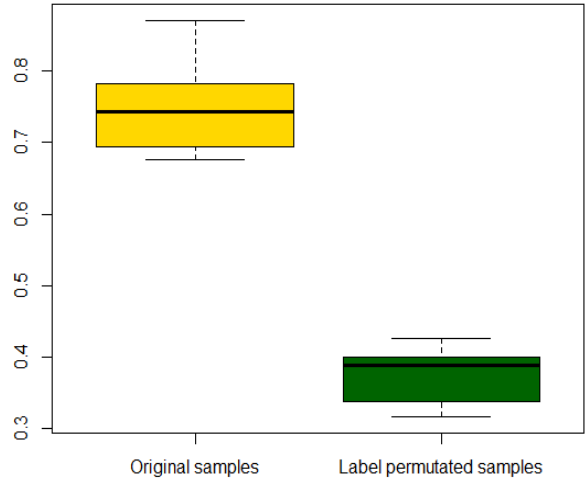

Figure 3 (b)

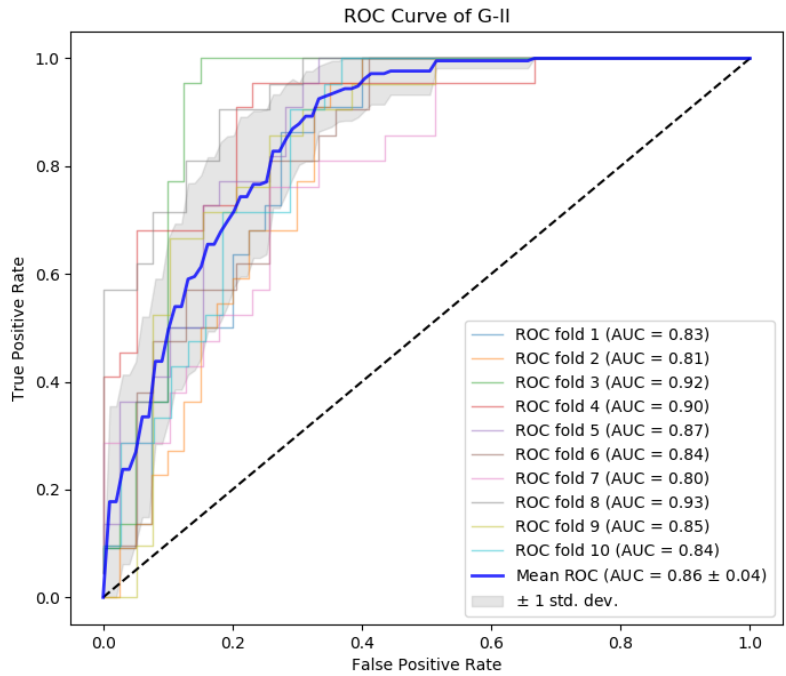

Figure 3 (c)

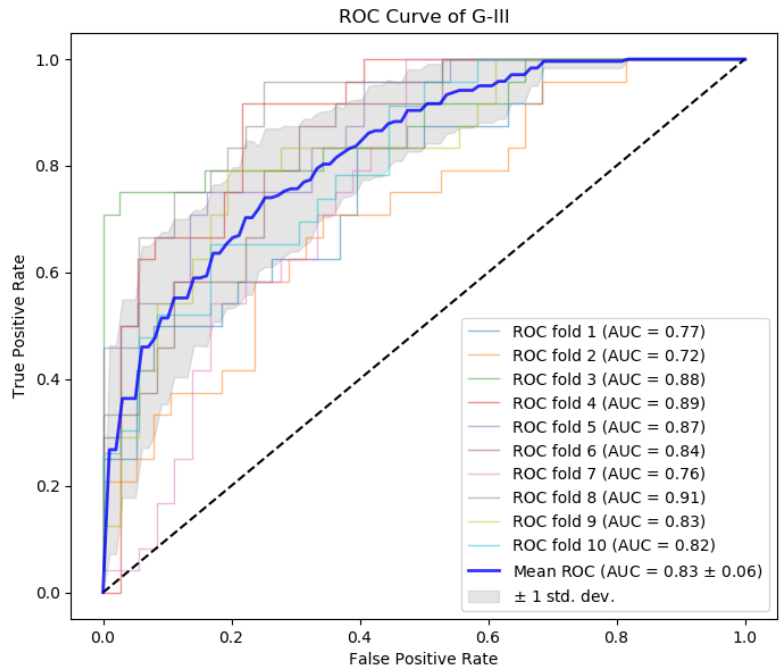

Figure 3 (d)

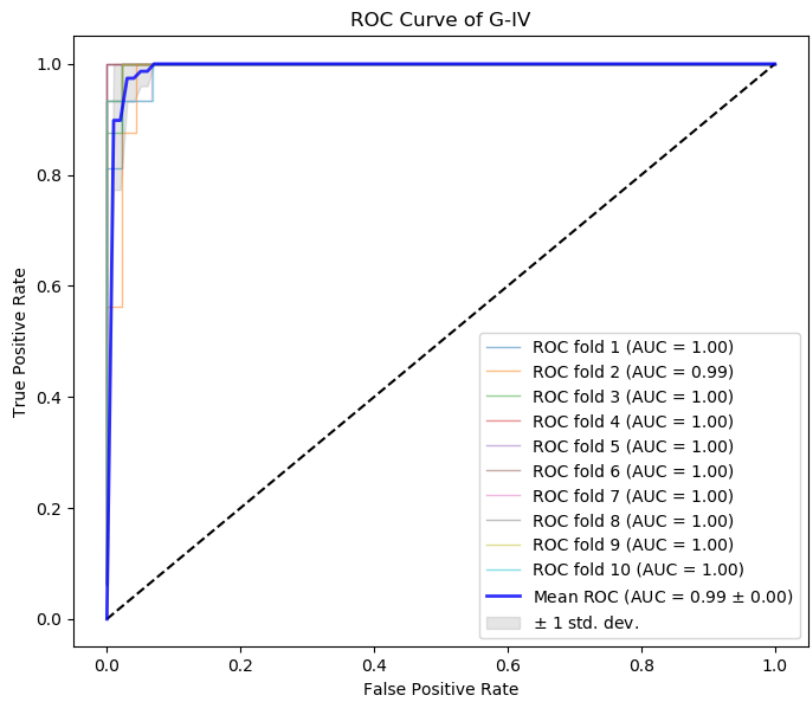

Figure 3 (e)

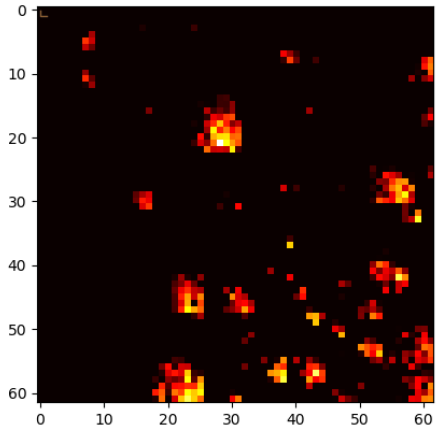

(a1)

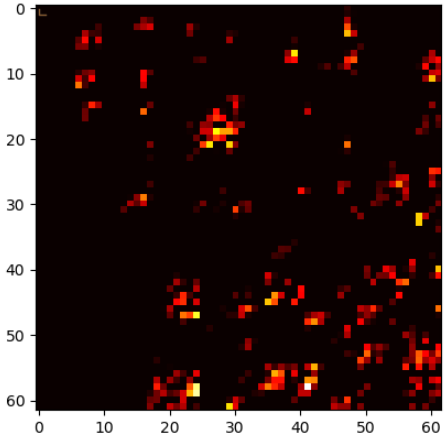

(a2)

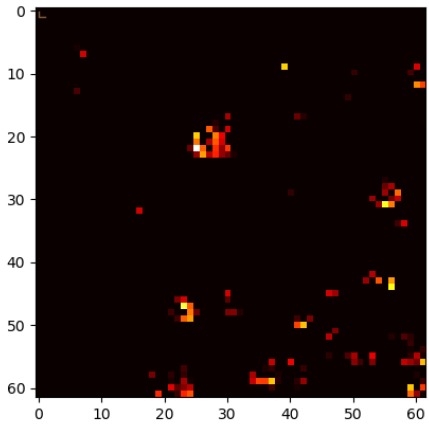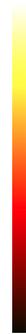

(a3)

Figure 4 (a1) (a2) (a3).

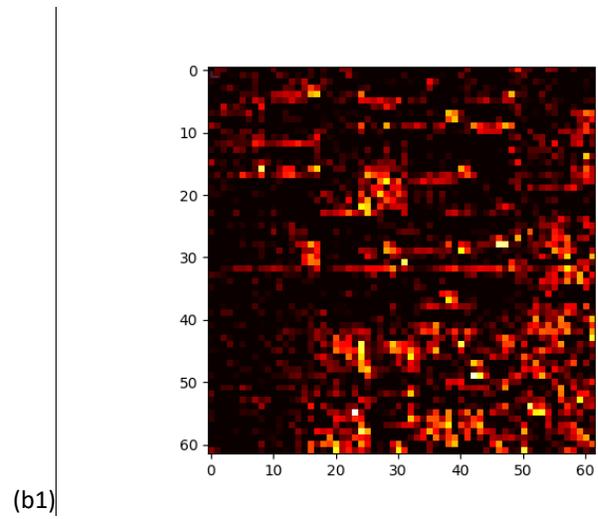

(b1)

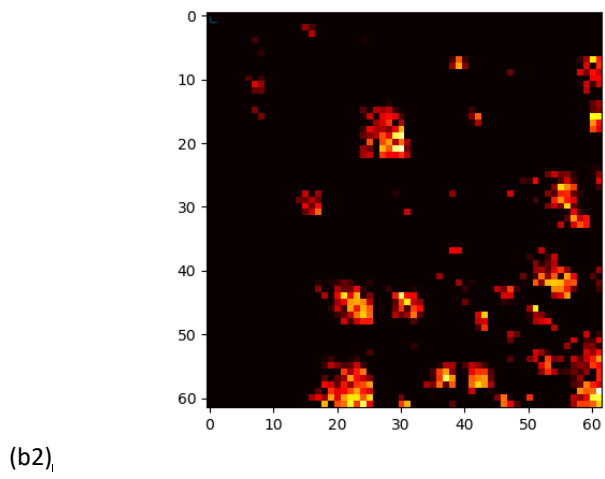

(b2)

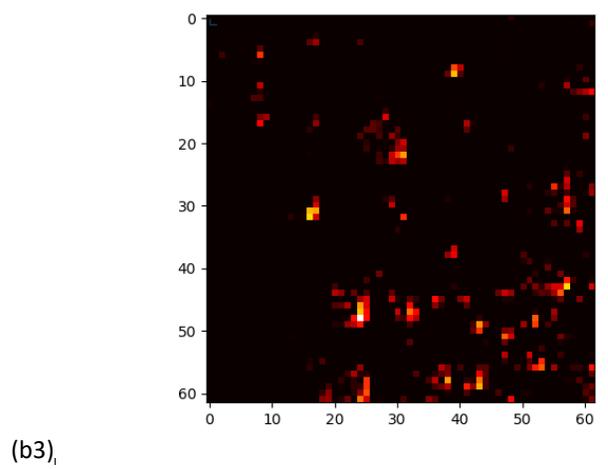
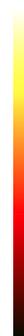

(b3)

Figure 4 (b1) (b2) (b3).

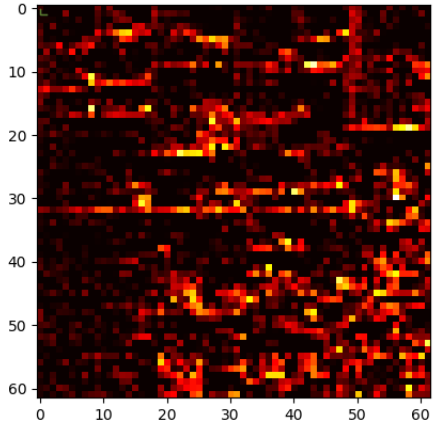

(c1)

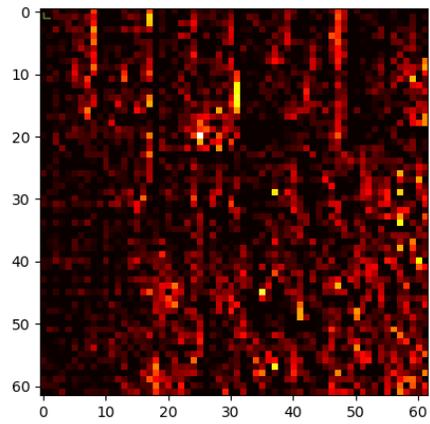

(c2)

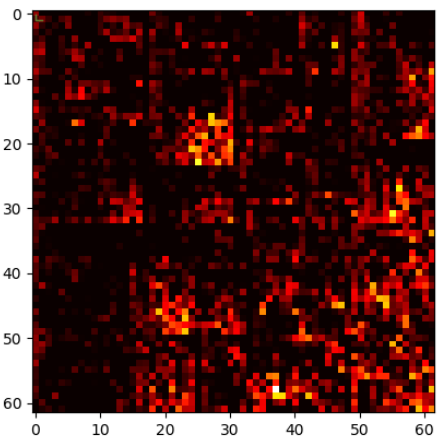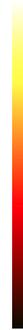

(c3)

Figure 4 (c1) (c2) (c3).

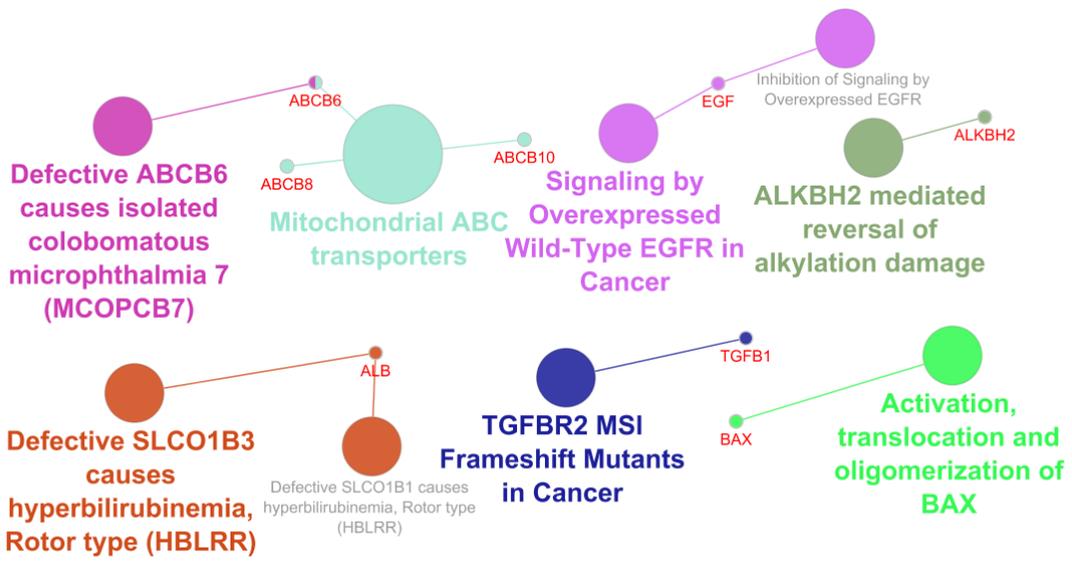

Figure 5

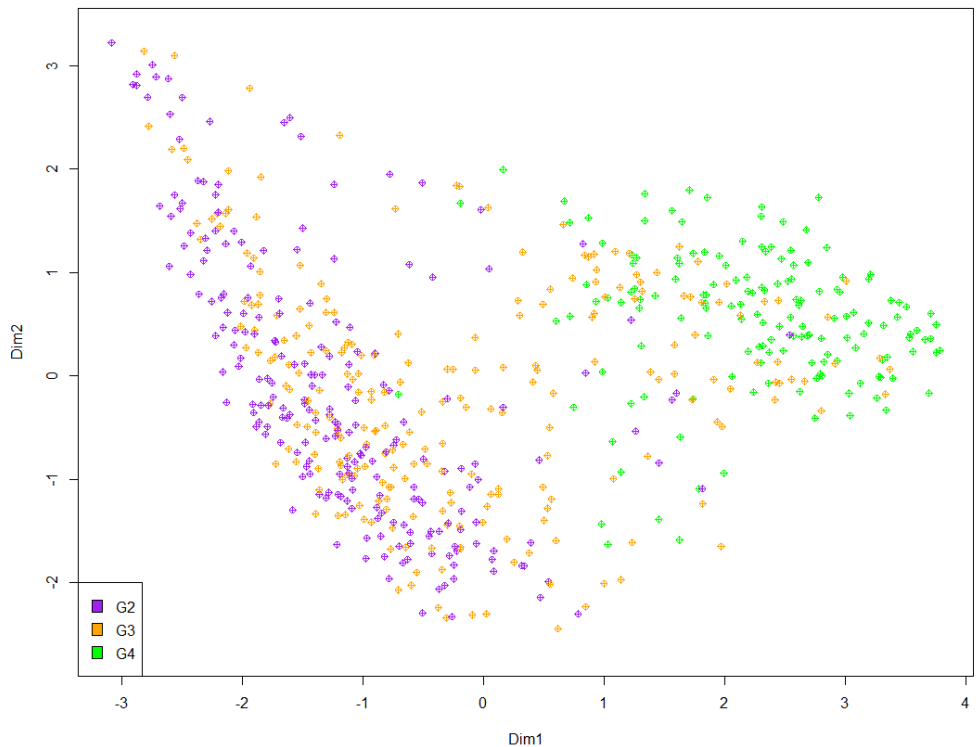

Figure 6 (a)

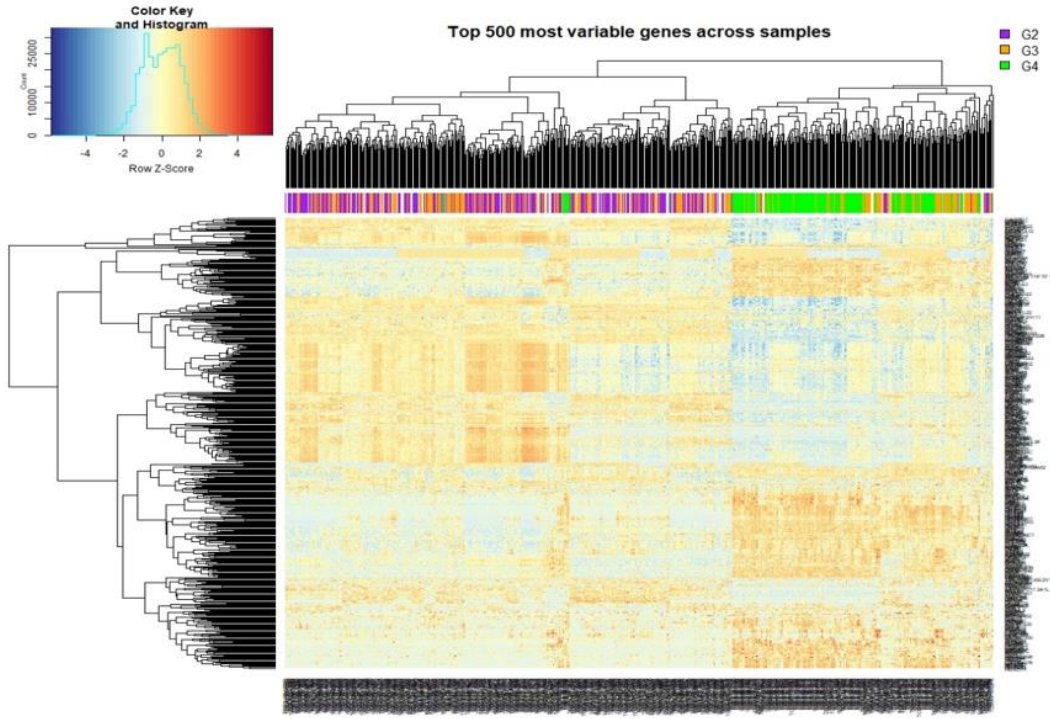

Figure 6 (b)

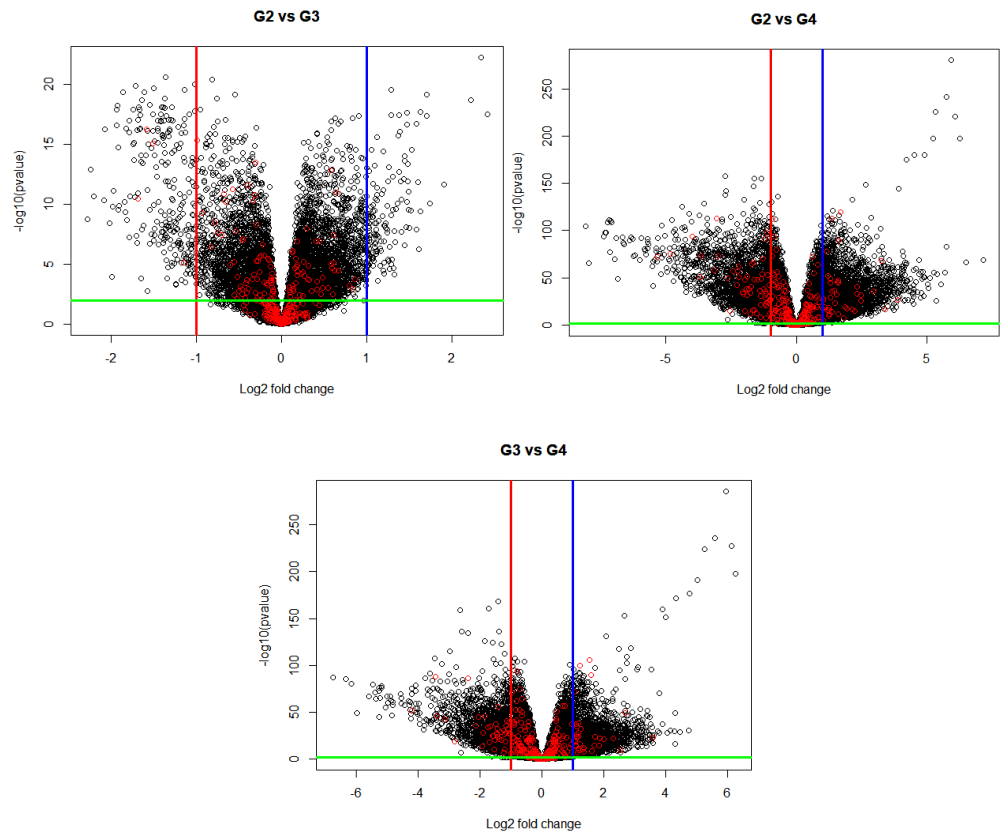

Figure 6 (c)

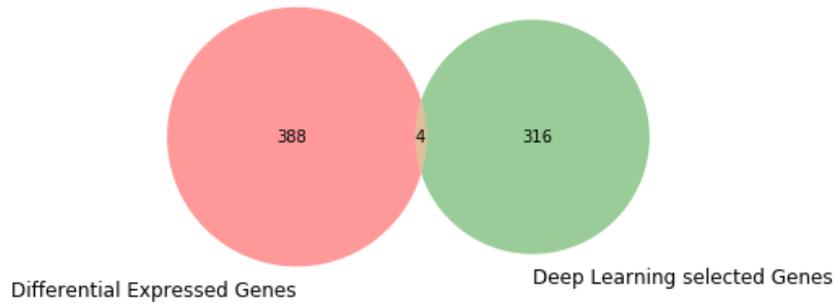

Figure 6 (d)

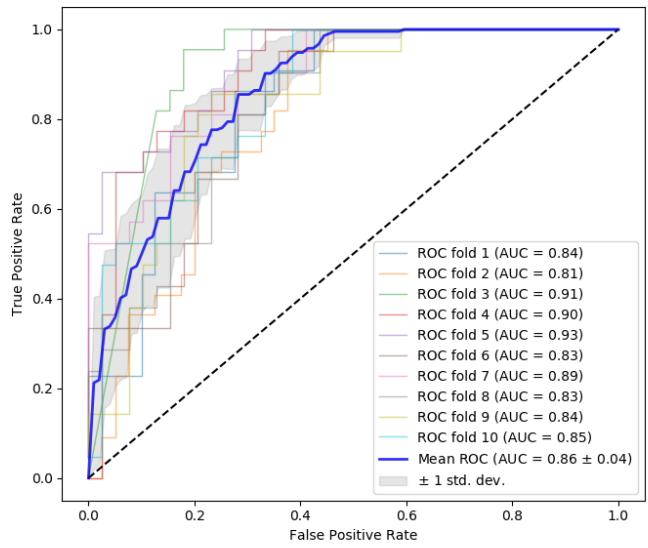

Figure 7 (a)

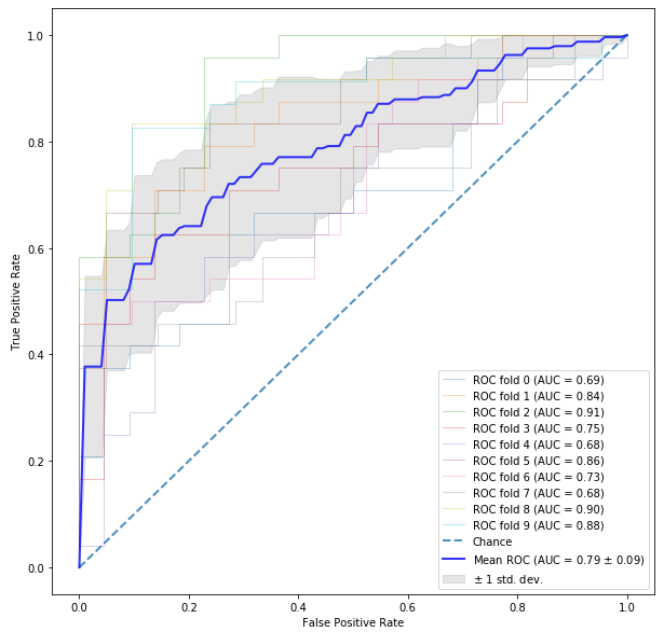

Figure 7 (b)

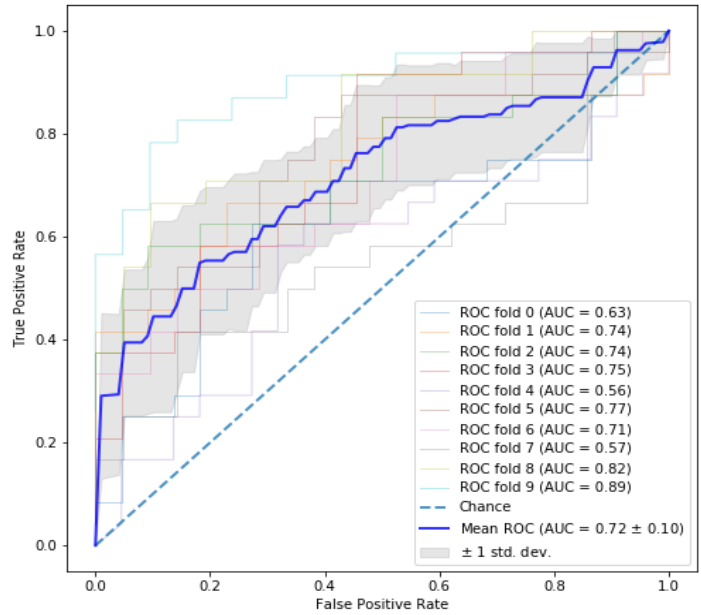

Figure 7 (c)

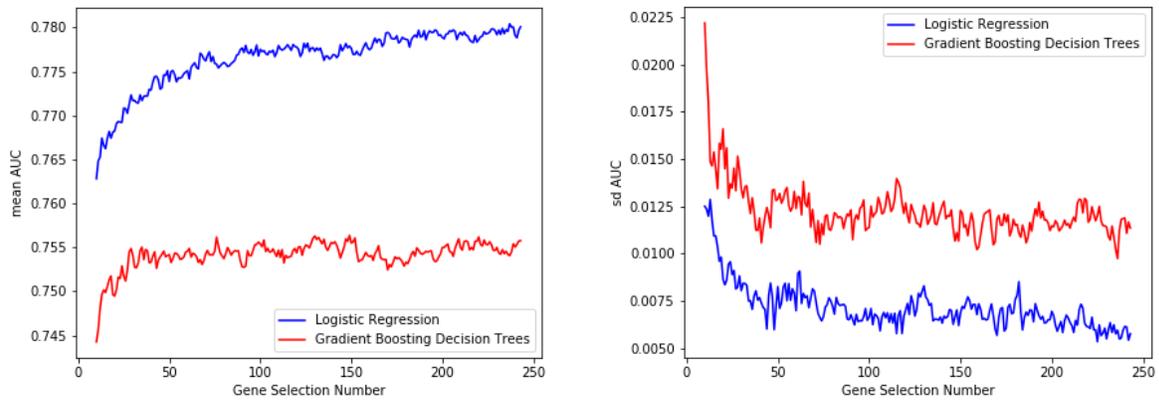

Figure 7 (d)